\pdfoutput=1

\documentclass[11pt]{article}

\usepackage[]{ACL2023}

\usepackage{times}
\usepackage{latexsym}
\usepackage{graphicx} 
\usepackage{hyperref}

\usepackage[T1]{fontenc}

\usepackage[utf8]{inputenc}

\usepackage{microtype}

\usepackage{inconsolata}

\title{Semi-supervised News Discourse Profiling with Contrastive Learning}

\author{Ming Li \\
  Texas A\&M University \\
  \texttt{liming@tamu.edu} \\\And
  Ruihong Huang \\
  Texas A\&M University \\
  \texttt{huangrh@cse.tamu.edu} \\}

\begin{document}
\maketitle
\begin{abstract}
News Discourse Profiling seeks to scrutinize the event-related role of each sentence in a news article and has been proven useful across various downstream applications. Specifically, within the context of a given news discourse, each sentence is assigned to a pre-defined category contingent upon its depiction of the news event structure. However, existing approaches suffer from an inadequacy of available human-annotated data, due to the laborious and time-intensive nature of generating discourse-level annotations. 
In this paper, we present a novel approach, denoted as Intra-document Contrastive Learning with Distillation (ICLD), for addressing the news discourse profiling task, capitalizing on its unique structural characteristics. Notably, we are the first to apply a semi-supervised methodology within this task paradigm, 
and evaluation demonstrates the effectiveness of the presented approach. Codes, models, and data will be available. \footnote{\href{https://github.com/MingLiiii/ICLD}{https://github.com/MingLiiii/ICLD}} 
\end{abstract}

\section{Introduction}

News discourse profiling \cite{choubey-etal-2020-discourse} is a specialized task aimed at comprehensively analyzing the structural aspects of news articles and effectively categorizing each sentence based on its contextual depiction of news events. Therefore, this is a document-level task with sentence-level predictions  \cite{li2022survey}, which has been proven useful in several downstream tasks, including text simplification \cite{zhang-etal-2022-predicting}, media bias analysis \cite{lei2022sentence}, event coreference resolution \cite{choubey-etal-2020-discourse}, RST-style Discourse Parsing \cite{li2023rststyle} and temporal dependency graph building \cite{choubey-huang-2022-modeling}. 

\begin{figure}[t]
\centering 
\includegraphics[width=0.5\textwidth]{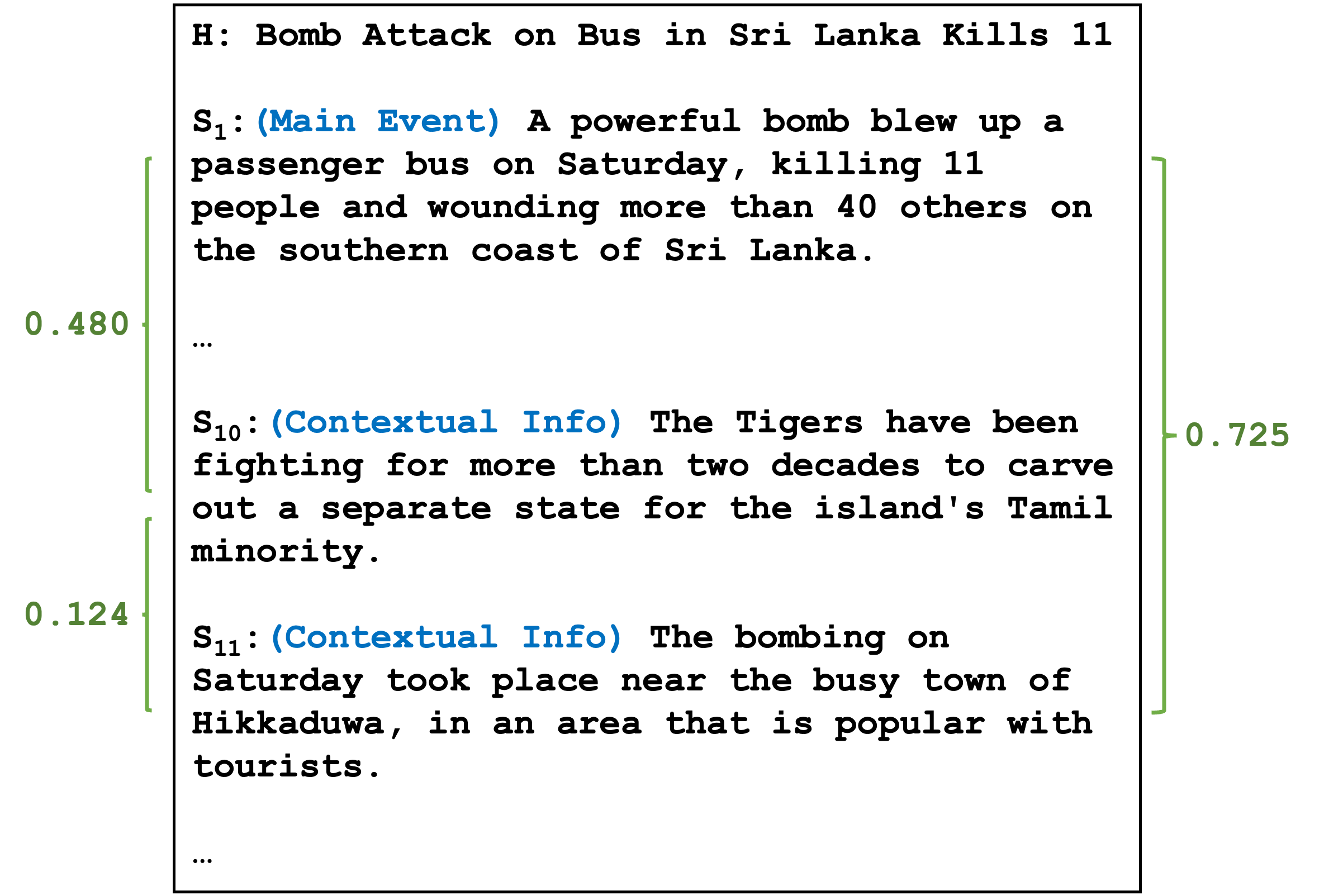} 
\caption{
An example of news articles in the human-annotated data. 
\textit{H} represents the headline of the news and \textit{S$_i$} represents the $i$th sentence in the news. 
In parenthesis (blue) are news discourse profiling labels assigned to the respective sentences. 
On the left and right sides (green), we provide the cosine similarity values derived from the sentence embeddings generated by Google's Universal Sentence Encoder. 
In this example, \textit{S$_1$} and \textit{S$_{11}$} are semantically similar ($0.725$) but have different labels. 
However, \textit{S$_{10}$} and \textit{S$_{11}$} are in the same category even though they are not semantically similar ($0.124$). 
} 
\label{example} 
\end{figure} 

Nevertheless, as a discourse-level task, the process of creating annotations entails a substantial investment of time and labor. 
The absence of human-annotated data poses a significant obstacle to the practical implementation of news discourse profiling, despite the relatively straightforward acquisition of unlabeled news articles. Building upon the aforementioned rationale, our impetus resides in introducing more unlabeled data for training purposes and formulating a semi-supervised methodology tailored to this particular task structure. 

Contrastive learning \cite{zhang-etal-2022-contrastive-data,le2020contrastive, albelwi2022survey} can effectively make use of unlabeled data and has been developed greatly recently which aims to learn effective representations of words, sentences, or discourses by pulling semantically close samples together and pushing away others \cite{gao-etal-2021-simcse}. 
A fundamental premise underlying contrastive learning is that the features acquired by encoders, via self-identification, encompass crucial information capable of not only distinguishing individual instances but also discerning disparities across different classes \cite{nce}. 
However, in news discourse profiling, the classification of each sentence is more profoundly influenced by the collective discourse structure and its interrelationships with other sentences, rather than relying solely on the inherent semantic meanings of individual sentences, which poses difficulties in designing a contrastive learning methodology for this task. 

As depicted in Figure \ref{example}, we present an illustrative instance selected from human-annotated datasets. \textit{H} denotes the news headline, while \textit{S$_i$} represents the $i$-th sentence within the news\footnote{The ordering of sentences is important in analyzing the event structure of news articles.}. In parenthesis (blue) are news discourse profiling labels assigned to the respective sentences. On the left and right sides (green), we provide the cosine similarity values derived from the sentence embeddings generated by Google's Universal Sentence Encoder \cite{yang-etal-2020-multilingual}. 

Although \textit{S$_{10}$} and \textit{S$_{11}$} serve similar functions by elucidating the terrorists' workplace and specifying the precise detonation site of the bomb, their similarity score is a mere $0.124$. Conversely, \textit{S$_{1}$} and \textit{S$_{11}$} exhibit a higher similarity score of $0.725$, despite their distinct narrative roles in conveying this news account. Consequently, it is evident that a substantial discrepancy exists between the assigned sentence categories and their underlying semantic significance, underscoring a pronounced misalignment.
In such a scenario, conventional sentence-level contrastive learning approaches prove inadequate for enhancing news discourse profiling, primarily due to their emphasis on capturing sentence-level semantic meanings. Furthermore, standalone sentences devoid of contextual information lack the capacity to effectively represent the intricate high-level event structures characterizing the entire discourse.

Building upon the aforementioned discussions, our objective is to establish an embedding space that not only captures semantic similarities but also incorporates the underlying event structure. 
Diverging from conventional contrastive learning methods that construct instance pairs through self-supervision, our approach operates in a semi-supervised manner. 
Thus we present a novel semi-supervised approach for news discourse profiling, termed Intra-document Contrastive Learning with Distillation (ICLD), specifically designed for this discourse-level task. 
In our proposed method, we employ a teacher model to predict silver labels for unlabeled news articles that have not been previously seen. These predicted labels act as guiding signals for the construction of positive and negative sentence pairs within each document, facilitating the contrastive learning process. 
Furthermore, our method incorporates intra-document contrastive learning along with an additional knowledge distillation component. This serves two purposes: firstly, to ensure the interaction between the target sentence and its contextual surroundings, and secondly, to further prevent the collapse of the contrastive aspect into simply learning the semantics similarities of individual sentences.

Extensive experimental evaluations have been conducted, confirming the efficacy of our proposed method. By incorporating a larger volume of readily accessible unlabeled news articles, we achieve a significant improvement in news discourse profiling performance. Notably, to the best of our knowledge, we are the first to address this particular task structure and propose a semi-supervised methodology to tackle it.

\section{Related Work}

\subsection{Contrastive Learning}
Recently the technique contrastive learning has been widely used in unsupervised and self-supervised learning, which greatly improved the performance of both visual and language representation \cite{simclr, moco, simcse, wu2020clear, zhang2021supporting, janson2021semantic}. 
It learns the data representation by pushing away negative samples and pulling close the positive samples where InfoNCE \cite{nce} objective is mostly used. 
Ideally, this would update the encoder to carry enough information for both sample identification and downstream classification. 

After achieving great success in computer vision tasks \cite{simclr, moco}, contrastive learning methods are then applied to the Natural language processing (NLP) area for sentence representation learning \cite{simcse, wu2020clear, zhang2021supporting, janson2021semantic}. 
One of the main methodological differences among these works is the method of data augmentation to generate positive pairs. 
{CLEAR} \cite{wu2020clear} utilizes word deletion, span deletion, reordering, and substitution for data augmentation. 
It calculates objectives on both the token level and the sentence level. 
{DeCLUTR} \cite{giorgi2020declutr} treats sentences from the same documents as positive pairs, while sentences from different documents as negative pairs. 
{SimSCE} \cite{simcse} utilizes the dropout in the pretrained word encoder and is proven to be an efficient way of augmentation. 
Leveraging the foundational concepts of SimCSE, a plethora of subsequent research endeavors have sought to enhance this framework through the incorporation of advanced auxiliary training objectives \cite{chuang-etal-2022-diffcse, nishikawa-etal-2022-ease, zhou2023learning, wu-etal-2022-infocse}, and \cite{chanchani-huang-2023-composition} recently proposes maximizing alignment between texts and composition of their phrasal constituents.

\subsection{Knowledge Distillation}

Knowledge distillation is first proposed by \cite{kd} for model compression by minimizing the KL divergence between the output distributions of the teacher model and the student model. 

In NLP tasks, large pretrained language models have achieved remarkable performance \cite{devlin-etal-2019-bert, qiu2020pre, alkhamissi2022review}. 
Knowledge distillation is one way to retain comparable performance as large models with relatively compact models. 
\cite{Unikd} effectively compresses models like BERT-base to BERT-4. 
\cite{plmKT} comes up with a framework for finetuning a domain-specific pretrained large language model as a teacher, then uses activation boundary distillation to teach domain knowledge to another language model. 
\cite{multi-granularity} compares the effect of knowledge in three different levels: token level, span level, and sample level among which the sample level maintains most of the knowledge. 
\cite{sparse} pretrains and finetunes a teacher model without pruning, then progressively replaces layers of the teacher model with the student model learned by knowledge distillation, which mitigates the overfitting in finetuning pretrained language models.

\section{The Semi-supervised Method}
Our semi-supervised approach consists of two learning phases, the first phase of Intra-document Contrastive Learning with Distillation (ICLD) exclusively utilizes unlabeled news articles and the second phase brings back human annotations to better calibrate the model. 
\begin{figure}[t]
\centering 
\includegraphics[width=0.48\textwidth]{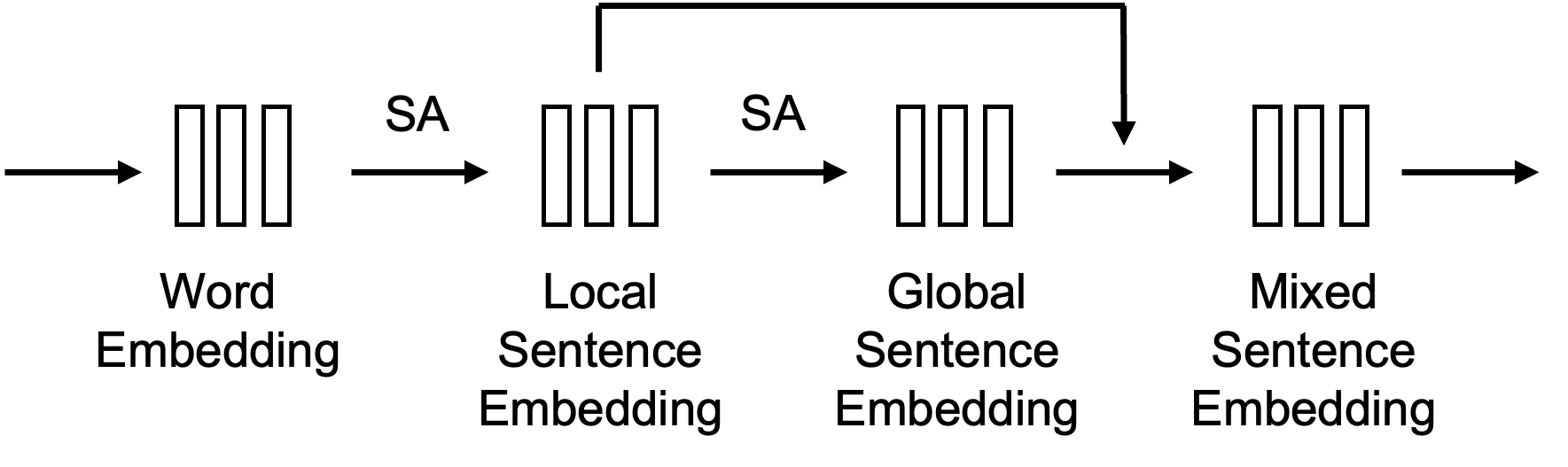} 
\caption{
A brief illustration of our model structure following \citet{li2022less}. 
\textit{SA} represents the self-attention module. 
Upon receiving an input discourse, initial word embeddings are generated using a language model. Subsequently, two self-attention modules are employed to obtain both local and global sentence embeddings. The mixed sentence embeddings are derived by adding the local and global sentence embeddings, followed by the integration of a fully connected layer. Finally, a classification layer is applied to yield the final prediction. 
The intra-document contrastive learning phase will be implemented upon mixed sentence embeddings. 
} 
\label{struct} 
\end{figure}

\subsection{Model Structure}

Within the context of the semi-supervised framework, the teacher model encounters a substantial volume of unlabeled news articles that may exhibit diverse distributions distinct from the human-annotated training set. Consequently, to offer more reliable guidance, the teacher model must possess strong generalization capabilities, ensuring the accuracy of the generated labels for unseen data instances. In light of these considerations, we opt for LiMNet \cite{li2022less}, incorporating the robust T5 large language model \cite{JMLR:v21:20-074}, as our selected teacher model due to its commendable performance and generalization abilities. 

Our student models adopt the same structural framework as LiMNet, leveraging small language models such as Longformer \cite{beltagy2020longformer} as the default choice. During training, the weights of these student models are iteratively updated. It is worth noting that the teacher model is solely trained using the original annotated data, adhering to the identical configuration outlined in its original paper. Subsequently, the automatically collected unlabeled news articles are fed into this well-trained teacher model, enabling the derivation of probability distributions across different sentence categories.

Figure \ref{struct} illustrates the simplified model architecture of LiMNet, where two self-attention modules \cite{bahdanau2014neural, NIPS2015_1068c6e4} are utilized. The first self-attention module focuses on capturing interactions among word embeddings within the given sentence, thereby producing a localized representation as the local sentence embedding. The second self-attention module leverages the interaction between the specific sentence embedding and the contextual word embeddings to derive a comprehensive global sentence embedding. Finally, a fully connected layer is employed for the purpose of final classification.

\begin{figure}[t]
\centering 
\includegraphics[width=0.48\textwidth]{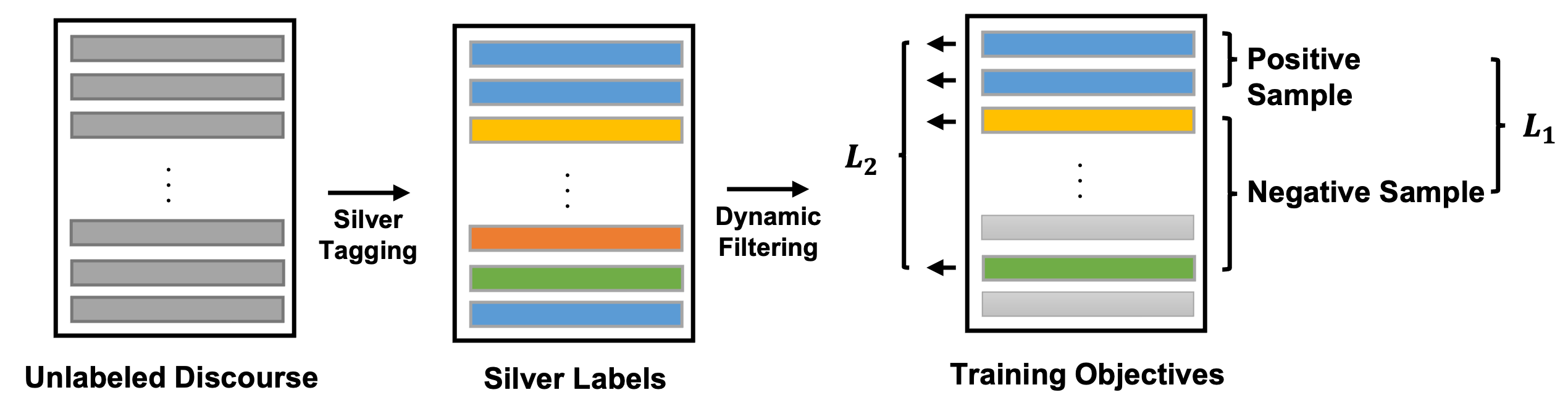} 
\caption{
An overview of our proposed method. For a given unlabeled discourse, the teacher model is first utilized to tag the sentences of the discourse to obtain the silver labels, where different colors denote different classes. Then dynamic filtering is implemented to filter those sentences with relatively low confidence, the gray color means the sentences are neglected. Then, sentences in the same category are randomly sampled as positive samples while sentences with different categories are randomly sampled as negative ones, on which is the contrastive training objective $L_1$ calculated. In the meantime, an extra distillation phase $L_2$ is implemented on unfiltered sentences to avoid the collapse solution. 
} 
\label{ICLD} 
\end{figure} 

\subsection{Phase One: Intra-document Contrastive Learning with Distillation (ICLD)}
During the ICLD phase, exclusively unlabeled news articles are utilized, figure \ref{ICLD} illustrates the workflow of intra-document contrastive Learning with distillation (ICLD). 

\vspace{.05in}
\noindent
\textbf{Random Sentence Filtering} 
Due to the potential distribution shift between the human-annotated data and the newly acquired unlabeled data, strictly adhering to the generated silver labels may lead to suboptimal outcomes. Moreover, the imperfections of the teacher model could be transmitted to the student models through the inclusion of noisy silver labels. To address this challenge, we propose a solution involving the random filtering of sentences exhibiting relatively lower confidence.

Considering the variability in confidence distributions across different categories, it is impractical to manually establish a specific static threshold for each category. Instead, we utilize a more flexible approach that leverages the intrinsic characteristics of the teacher models and the newly collected articles. Specifically, we adopt the $k$-th percentile approach for determining flexible thresholds, based on the silver label confidences estimated by the teacher model. These thresholds are dynamically adjusted as the teacher model or unseen data undergo modifications. Consequently, sentences with confidences lower than their respective thresholds are subjected to filtering with a probability of $0.5$ during each epoch. This stochastic filtering approach is employed due to the inherent uncertainty associated with low-confidence sentences, as their definitive correctness cannot be ascertained in the absence of golden labels. By employing adaptable thresholds, we mitigate the reliance on predefined confidence thresholds for individual categories, allowing for a more tailored and nuanced filtering process.

\vspace{.05in}
\noindent
\textbf{Intra-document Contrastive Learning (ICL)} 

The generation of positive pairs for contrastive learning follows a specific procedure: under the supervision of silver labels, unfiltered sentence pairs of the same category within each document are randomly sampled without replacement until no sentences remain belonging to the same category. Then, negative pairs are randomly sampled without replacement, from the remaining sentences where no sentence pair shares the same silver label until no additional pairs can be generated. Specifically, once the silver labels are procured, for each label with more than two associated sentences, we randomly select two sentences. This process continues until just one or no sentence remains for that label, constituting our positive samples. For the remaining sentences, we repeat a similar sampling of two sentences at a time to form negative pairs until only one sentence or none remains in the document. This sampling process guarantees diverse positive and negative pairs, facilitating effective contrastive learning within the intra-document context.

Following the establishment of positive and negative pairs, the contrastive learning constraint is imposed on the mixed sentence embeddings derived from the input discourse. Notably, the mixed sentence embeddings encompass crucial contextual information essential for comprehending the news event content, distinguishing them from lower-level sentence embeddings. Thus, the contrastive learning process focuses on leveraging contextual embeddings to enhance the discriminative ability and semantic understanding of the news event representations.
Cosine similarity is used as the measurement of similarity between two embeddings, and the contrastive learning constraint is formulated as:
\begin{eqnarray}
     L_1 = -log\frac{m\cdot\sum_{i=1}^{n}{e^{cos(g_i,g^-_i)}/\tau}}{n\cdot\sum_{j=1}^{m}{e^{cos(g_j,g^+_j)}/\tau}+\beta}
\end{eqnarray}
where $g_i$ and $g^-_i$ represent global sentence embeddings in negative pair, $g_j$ and $g^+_j$ represent sentence embeddings in positive pair. 
$n$ is the number of positive pairs and $m$ is the number of negative pairs in this discourse. 
$\tau$ represents the temperature rate which is in the range of $0$ to $1$. 
$\beta$ represents a small value to avoid a division by zero and is set to $1e-6$ by default. 

\vspace{.05in}
\noindent
\textbf{Knowledge Distillation} 

In contrast to tasks that primarily emphasize the semantic meanings of individual sentences, news discourse profiling directs its attention toward the collective representation of sentences in describing news events. To avoid the potential collapse of task-specific intra-document contrastive learning towards standard contrastive learning, which prioritizes the semantic meanings of individual sentences, the incorporation of explicit guidance and simultaneous distillation becomes imperative.

To address this challenge, silver labels simultaneously serve as direct training guidance for the student model. This facilitates a more explicit and informed learning process. Concurrently, flexible threshold-based random filtering is applied to eliminate low-confidence sentences, ensuring the optimization of the student model with reliable and informative training instances. By leveraging these measures, this task-specific intra-document contrastive learning remains focused on capturing the interdependencies and contextual cues within the news discourse, promoting a more accurate and contextually rich news discourse profiling.

Different from training with human-annotated data where cross-entropy is used, we choose to minimize the mean square error (MSE) of probability distribution between the silver labels and the student model, which can be formulated as: 
\begin{eqnarray}
      L_2 = \frac{1}{l} \sum_{i=1}^{l} (y_i - \hat{y}_i)^2 
\end{eqnarray}
where $\hat{y}_i$ represents the probability distribution from the student model, and $y_i$ represents the silver labels generated by the teacher model. 
$l$ is the number of unfiltered sentences. 

Contrastive learning projects sentence embeddings to the desired space while knowledge distillation aims to build mappings between sentence representations and discourse role classes, therefore, balancing these two objectives can slow down classifier formation.
Empirically, we found that it facilitates training to enable both contrastive learning and knowledge distillation for the first few training epochs and then continues to train for more epochs with knowledge distillation as the only learning objective. When both intra-document contrastive Learning and knowledge distillation are enabled, the overall learning objective is the summation of $L_1$ and $L_2$. 

\subsection{Phase Two: Final Finetuning}
After training on unlabeled data in the ICLD phase, the human-annotated golden data are utilized to better calibrate the model. 
The final model finetuning will use cross-entropy loss as the objective:
\begin{eqnarray}
      L_3 = -\sum_{c=1}^{C} y_c \log(\hat{y}_c) 
\end{eqnarray}
where $\hat{y}$ represents the probability distribution from the student model, and $y$ represents the human-annotated gold labels. $C$ represents the number of classes. Overall, $L_1$ and $L_2$ are utilized upon unlabeled data, and $L_3$ is utilized upon human-annotated data.

\section{Evaluation}

\subsection{Dataset}

\vspace{.05in}
\noindent
\textbf{Labeled data} 

The NewsDiscourse dataset \cite{choubey-etal-2020-discourse} we use is designed for the task of News Discourse Profiling, which consists of $802$ news articles ($18,155$ sentences). 
These news discourses are sampled from three news sources including NYT, Xinhua and Reuters and they are in four domains including business, crime, disaster, and politics. 
Each sentence in this corpus is labeled with one of eight content types\footnote{The eight content types are grouped into three categories: Main Content, Context-informing Content, and Additional Supportive Content. 
In Main Content, there are two fine-grained categories: \textit{Main Event} which introduces the most important event relating to the major subjects of news discourse, and \textit{Consequence} which represents content that is triggered by the main news event. 
In Context-informing Content, there are two fine-grained categories: \textit{Previous Event}  which precedes the
the main event and now acts as possible causes or preconditions for the main event, and \textit{Current Context} which covers all the context informing the main event. 
In Additional Supportive Content, there are four fine-grained categories: \textit{Historical Event} which represents events that precede the main event in months or
years, \textit{Anecdotal Event} which represents unverified events of a person or situation, \textit{Evaluation} which represents opinionated contents including reactions from immediate participants, experts, known personalities, as well as journalists or news sources and \textit{Expectation} represents speculations and projected consequences. } representing what role it plays in reporting a news story or the "None" class, following the news content schemata proposed by Van Dijk \cite{van1985structures, van1988news}. 
Following \citet{choubey-huang-2021-profiling-news}, we use $502$ documents for training, $100$ documents for validation, and $200$ documents for testing. 
All the models are evaluated by calculating the micro F1 and macro Precision, Recall, and F1 scores are implemented form the scikit-learn library \cite{pedregosa2011scikit}.

\vspace{.05in}
\noindent
\textbf{Unlabeled data} 

To simulate the real-world application scenario where the data distribution can largely vary from the existing human-annotated data, we deliberately avoid following the same source and domains from the existing data. 
Unlabeled news articles are collected from CNN with a variety of domains including business, entertainment, health, politics, sports, style, travel, and world. 
These unlabeled data contain $10,337$ news articles with $135,057$ sentences and all of these data will be used in our method. 
The improvement made by our method using these data proves that our method is effective even when the unlabeled data have different distributions than the original data.

\subsection{Implementation Details}

All experiments are implemented in the PyTorch platform \cite{paszke2019pytorch} with one NVIDIA A100 graphic card. 

All pre-trained language models used in this paper are implemented from \emph{huggingface} \cite{wolf2019huggingface}. 
Our teacher model utilizes \textit{large} version of T5 as the pretrained language model, while all our student models use the \textit{base} version of pretrained language models with the output dimension of $768$. 
The parameters of the language model in the teacher model are fixed all the time including its training phase, and the parameters of the language model in the student models are not fixed and are updated during training.

Our models are trained using Adam optimizer \cite{kingma2014adam} with the hyper-parameters betas=[$0.9$, $0.999$], eps=$1e-8$ and the learning rate is set to $5e-6$ for $25$ epochs. 
The dropout rate \cite{JMLR:v15:srivastava14a} is set to $0.5$. 
Intra-document contrastive learning with distillation (ICLD) is applied in the first $3$ epochs, where both $L_1$ and $L_2$ are utilized. Knowledge distillation from unlabeled data will continue for another 12 epochs with only $L_2$ as the learning objective. 
In the first 15 epochs, only unlabeled data with silver labels are utilized. Then, we will continue to train for $10$ epochs using only the original well-labeled data and the learning objective of $L_3$. 
We use the $50$ percentile as the threshold and filter sentences with the probability of $0.5$ by default. 
$\tau$ in contrastive learning constraint is set to $1$ by default.

\subsection{Ablation Study}

\begin{table}[h]
\centering
\scalebox{0.75}{\begin{tabular}{l|c|c|c|c}
\hline
 & \multicolumn{3}{|c|}{\textbf{Macro}} & \textbf{Micro}\\
\hline
\ & Precision & Recall & F1 & F1 \\ \hline
\  The full model & $\bf 67.9$ & $ \bf 68.8 $ & $ \bf 68.3$ & $ \bf 71.0 $\\ 
\  {\it w/o ICL} & $66.9$ & $68.3$ & $67.5$ & $70.1 $\\
\  {\it w/o Distillation} & $63.2$ & $65.0$ & $63.6$ & $66.5 $\\ \hline
\  {\it Positive Only} & $67.0 $ & $ 68.0$ & $67.4$ & $70.5$\\ 
\  {\it Negative Only} & $67.6$ & $69.2$ & $\bf 68.3$ & $70.8$\\ 
\hline
\end{tabular}}
\caption{\label{tbl:ablation_ecl}
Ablation experiments of our ICLD method.  
\textit{w/o ICL} represents the model where intra-document contrastive learning is not utilized, which becomes a standard knowledge distillation method. 
\textit{w/o Distillation} represents the model where no extra distillation is utilized simultaneously with intra-document contrastive learning. 
\textit{Positive Only} and \textit{Negative Only} represent experiments where only positive or negative pairs are sampled and calculated in the contrastive constraint. 
}
\end{table}

To evaluate the individual contributions of different components in our proposed Intra-document Contrastive Learning with Distillation approach, an ablation study is conducted, and the results are presented in Table \ref{tbl:ablation_ecl}. The model labeled as \textit{Without ICL} refers to the variant where intra-document contrastive learning is not incorporated. In this configuration, the model is trained solely with silver and real labels using the same configuration as our final ICLD model, which essentially functions as a standard knowledge distillation method. Comparing the performance of this variant with the complete ICLD model, we observe improvements across all metrics, demonstrating the efficacy of our intra-document contrastive learning component.

On the other hand, the model denoted as \textit{Without Distillation} represents the variant where no distillation process is executed concurrently with intra-document contrastive learning. Notably, this configuration exhibits a significant decline in performance. The absence of the distillation component leads to the collapse of intra-document contrastive learning into standard contrastive learning, where the classification of instance pairs primarily relies on their semantic meanings rather than task-specific discourse structure information. This comparative analysis underscores the indispensable nature of the additional distillation process, which furnishes the necessary guidance for the model to acquire task-specific structural information.
Overall, the ablation study highlights the importance of both intra-document contrastive learning and distillation in our approach, as they collectively contribute to the enhanced performance in capturing the intricate structure and nuances of news discourse profiling.

In addition, we present experimental results with respect to the sentence pair selection strategy. \textit{Sample Positive} and \textit{Sample Negative} represent experiments where only positive or negative pairs are sampled and calculated in the contrastive constraint. Comparing the performance of these variants, we find that using only positive pairs yields inferior results compared to using only negative pairs. This observation is reasonable as relying solely on positive pairs fails to establish a clear decision boundary and the negative samples play the dominant role in this contrastive learning phase.

\subsection{Effects of Random Filtering}

\begin{table}[t]
\centering
\scalebox{0.75}{\begin{tabular}{l|c|c|c|c}
\hline
 & \multicolumn{3}{|c|}{\textbf{Macro}} & \textbf{Micro}\\
\hline
\ & Precision & Recall & F1 & F1 \\ \hline
\  Probability of $0$ & $ 67.2$ & $ 68.3 $ & $ 67.6 $ & $ 70.5 $\\ 
\  Probability of $0.3$ & $ 67.4$ & $ 68.4 $ & $ 67.8 $ & $ 70.6 $\\
\  Probability of $0.5$ & $ \bf 67.9$ & $ 68.8 $ & $ \bf 68.3$ & $ \bf 71.0 $\\ 
\  Probability of $0.8$ & $67.8 $ & $ \bf 68.9$ & $68.2$ & $70.8$\\ 
\  Probability of $1.0$ & $67.8 $ & $ 68.3$ & $67.9$ & $\bf 71.0$\\ 
\hline
\end{tabular}}
\caption{\label{tbl:ablation_rf}
Extensive experiments with respect to the effects of random filtering. 
The probability value determines the likelihood of filtering sentences based on their confidence scores. When the probability is set to $0$, no filtering is applied, while a probability of $1.0$ indicates that every sentence with a confidence score below the threshold will be filtered.
}
\end{table}

In this section, we conducted experiments to investigate the effect of different random filtering probabilities ranging from $0$ to $1.0$, as presented in Table \ref{tbl:ablation_rf}. The default filtering threshold for these models was set to the $50$th percentile\footnote{The impact of using different confidence thresholds is discussed in Appendix A.}.
When the filtering probability is set to $0$, no sentences are filtered out. However, since there are no golden labels available for the newly collected news articles, the accuracy of the generated silver labels cannot be guaranteed. Without random filtering, the model might learn from potential noise in the silver labels, resulting in the lowest performance observed in the \textit{Probability of $0$} experiment. On the other hand, when the filtering probability is set to $1.0$, sentences with confidence scores lower than the threshold are entirely filtered out. However, filtering with a high probability is not optimal as it cannot be asserted that the low-confidence instances are definitively incorrect. Filtering out all of these low-confidence samples directly eliminates the possibility for the model to learn from them. Therefore, we chose a filtering probability of $0.5$ as the default setting.

\subsection{Comparison with Baselines}

\begin{table}[t]
\centering
\scalebox{0.75}{\begin{tabular}{l|c|c|c|c}
\hline
 & \multicolumn{3}{|c|}{\textbf{Macro}} & \textbf{Micro}\\
\hline
\ & Precision & Recall & F1 & F1 \\ \hline
\  Longformer (baseline)& $ 66.2$ & $ 62.3$ & $ 63.4 $ & $ 68.7 $\\ 
\  Longformer (ICLD)& $ \bf 67.9$ & $ \bf68.8 $ & $ \bf 68.3$ & $ \bf 71.0 $ \\ \hline
\  RoBERTa (baseline)& $ 66.6 $ & $ 61.5 $ & $ 62.9 $ & $ 69.0 $\\ 
\  RoBERTa (ICLD)& $ 67.2 $ & $ 67.1 $ & $ 67.1 $ & $ 70.8 $\\ \hline
\end{tabular}}
\caption{\label{tbl:ablation_fr} 
Comparisons with baseline models. 
Baseline models are trained with human-annotated data only. 
}
\end{table}

In this section, we compare our ICLD model with baselines using only original human-annotated data. 
\textit{Longformer (baseline)} and \textit{Longformer (ICLD)} utilize pretrained Longformer \cite{beltagy2020longformer} (base version) implemented from \textit{huggingface} \cite{wolf2019huggingface}. 
Compared with the baseline model where only human-annotated data is utilized, our ICLD model improves the macro F1 score by $4.9$ percent and micro F1 score by $2.3$ percent. 
\textit{RoBERTa (baseline)} and \textit{RoBERTa (ICLD)} utilize pretrained RoBERTa \cite{Liu2019RoBERTaAR} (base version) implemented from \textit{huggingface} \cite{wolf2019huggingface}. 
Compared with the baseline model where only human-annotated data is utilized, our ICLD model improves the macro F1 score by $4.2$ percent and micro F1 score by $1.8$ percent. 
These comparisons verify the effectiveness of our proposed method, which improves news discourse profiling by a large margin. 

Furthermore, it is observed that models utilizing Longformer demonstrate superior performance compared to those utilizing RoBERTa. RoBERTa is specifically designed to handle inputs within a token limit of 512, whereas news articles often exceed this limit. Consequently, we partition lengthy articles into multiple segments before feeding them into RoBERTa. In this process, sentences belonging to different segments are unable to interact with each other, resulting in the absence of comprehensive global contextual information within their corresponding sentence embeddings. Consequently, the performance is adversely affected. On the contrary, Longformer is explicitly designed to handle long inputs, thereby circumventing this potential segmentation issue. All sentences within a discourse can be collectively modeled, resulting in enhanced performance.

Notably, the performance of these two baseline models is relatively comparable since the standard training process does not fully exploit contextual information. For instance, when predicting the category of the first sentence, it is likely that knowledge of the last sentence is unexploited. 
Consequently, document splitting has minimal impact on performance. However, in our proposed contrastive-based method, the first and last sentences may be randomly selected in one positive or negative pair. In such cases, the sentence embeddings of these two sentences need to be compared and updated directly. With Longformer, all sentences share a similar contextual environment and possess a holistic view of the entire document, thereby justifying the comparison and update process. However, if the document is split when employing RoBERTa, these two sentences might belong to distinct semantic environments, making the comparison unjustifiable.

\subsection{Comparison with Previous Methods}

\begin{table}[t]
\centering
\scalebox{0.75}{\begin{tabular}{l|c|c|c|c}
\hline
 & \multicolumn{3}{|c|}{\textbf{Macro}} & \textbf{Micro}\\
\hline
\ & Precision & Recall & F1 & F1 \\ \hline
\  \cite{choubey-etal-2020-discourse} & $ 56.9$ & $ 53.7$ & $ 54.4 $ & $ 60.9 $\\ 
\  \cite{choubey-huang-2021-profiling-news}& $ 58.7$ & $ 56.4 $ & $ 57.0 $ & $ 62.2$ \\
\  \cite{spangher-etal-2021-multitask}& $ -$ & $ - $ & $ 63.5 $ & $ 67.5$ \\
\  \cite{li2022less} & $ \bf 68.2$ & $ 63.9 $ & $ 65.6 $ & $ 69.7$ \\
\hline
\  Our ICLD model & $ 67.9$ & $ \bf68.8 $ & $ \bf 68.3$ & $ \bf 71.0 $\\ \hline
\end{tabular}}
\caption{\label{tbl:previous}
Comparison with previous methods. 
\citet{spangher-etal-2021-multitask} does not report their macro precision and recall scores. 
}
\end{table}

Table \ref{tbl:previous} shows the performances of previous methods. 
In contrast to previous approaches, our semi-supervised method leverages newly introduced unlabeled data, resulting in a substantial improvement in news discourse profiling performance.
\citet{choubey-huang-2021-profiling-news} utilizes sub-topic information to guide the embedding extraction in an actor-critic manner. 
\citet{spangher-etal-2021-multitask} improves news discourse profiling performance by utilizing the multitask training from several discourse datasets. \footnote{For the detailed description of datasets, please see the Appendix in \citet{spangher-etal-2021-multitask}.} 
\citet{li2022less} mainly focuses on alleviating overfitting for discourse-level tasks and the performance in the table is based on T5 language model \cite{JMLR:v21:20-074}, which serves as our teacher model. 
It is worth noting that our ICLD model with the \textit{base} versions of Longformer or RoBERTa surpasses the performance of \citet{li2022less} where the \textit{large} version of T5 language model is utilized. 
Considering the huge discrepancy in model sizes and the training corpus utilized for these language models, we assert that our proposed method exhibits a significant capability for this task. 

\section{Conclusion}

In this work, we introduce the Intra-document Contrastive Learning with Distillation (ICLD) method for news discourse profiling, which leverages unlabeled data to enhance performance. News discourse profiling is a discourse-level task where the prediction of each sentence is intricately linked to the overall event structure of the entire discourse, rather than solely relying on the semantic meaning of individual sentences. To the best of our knowledge, we are the first to address this unique task structure and propose a semi-supervised approach to tackle it effectively. The dataset we collected emulates real-world scenarios, encompassing potential domain shifts, and our method demonstrates robust performance in such settings, affirming the efficacy of our proposed approach.

\section*{Limitations}

In this task, it is observed that sentences with similar semantic meanings can be assigned to different categories. Therefore, our objective is to establish an embedding space that not only captures semantic similarities but also incorporates the underlying event structure. To achieve this, we designed a contrastive learning based approach. 
However, it is important to note that our contrastive learning method is not necessarily the optimal solution when compared to other possible semi-supervised methods. 
The primary motivation of this paper is to address the unique task structure and propose a semi-supervised method that is specifically designed for this task. We do not claim that our method is comprehensive or superior but only serves as an initial exploration of a semi-supervised approach tailored to this intriguing task structure.

\section*{Acknowledgements}
We thank the anonymous reviewers for their valuable feedback and input. We gratefully acknowledge support from the National Science Foundation (NSF) via the awards IIS-1942918 and IIS-2127746. Portions of this research were conducted with the advanced computing resources provided by Texas A\&M High-Performance Research Computing. 

\bibliography{anthology,custom}
\bibliographystyle{acl_natbib}

\appendix

\section{Effects of Filtering Threshold}

\begin{table}[h]
\centering
\scalebox{0.75}{\begin{tabular}{l|c|c|c|c}
\hline
 & \multicolumn{3}{|c|}{\textbf{Macro}} & \textbf{Micro}\\
\hline
\ & Precision & Recall & F1 & F1 \\ \hline
\  $10$-Percentile & $ 67.6$ & $ 67.6 $ & $ 67.5 $ & $ 70.5 $\\ 
\  $30$-Percentile & $ 67.6$ & $ 68.2 $ & $ 67.8 $ & $ 70.6 $\\
\  $50$-Percentile & $ \bf 67.9$ & $ \bf68.8 $ & $ \bf 68.3$ & $ \bf 71.0 $\\ 
\  $70$-Percentile & $ \bf 67.9 $ & $ 68.2$ & $68.0$ & $70.7$\\ 
\  $90$-Percentile & $66.0 $ & $ 67.9$ & $66.8$ & $ 70.1$\\ 
\hline
\end{tabular}}
\caption{\label{tbl:ablation_fr}
Ablation experiments with respect to the filtering threshold. 
}
\end{table}

In order to investigate the effects of the filtering threshold on the performance of our method, we analyze the results using different thresholds in Table \ref{tbl:ablation_fr}. The default filtering probability for these models is set to $0.5$. From the table, it can be observed that this hyperparameter has a negligible effect on the model performance.
However, when the filtering threshold is set to the $90$th percentile, indicating that $90\%$ of sentences will be randomly filtered, a significant decrease in performance is observed. This outcome is expected as a high percentile threshold results in a substantial reduction in the amount of available unlabeled data, which affects the model's learning capabilities.

\section{Effects of Amount of Extra Data}

\begin{table}[h]
\centering
\scalebox{0.75}{\begin{tabular}{l|c|c|c|c}
\hline
 & \multicolumn{3}{|c|}{\textbf{Macro}} & \textbf{Micro}\\
\hline
\ & Precision & Recall & F1 & F1 \\ \hline
\  0 & $ 66.2 $ & $ 62.3 $ & $ 63.4 $ & $ 68.7 $\\ 
\  500 & $ 64.2 $ & $ 63.4 $ & $ 63.6 $ & $ 67.9 $\\ 
\  1,000 & $ 66.3 $ & $ 64.7 $ & $ 65.3 $ & $ 69.9 $\\
\  2,000 & $ 68.0 $ & $ 64.7 $ & $ 65.5 $ & $ 69.8 $\\ 
\  3,000 & $ 67.5 $ & $ 65.2 $ & $ 66.0 $ & $ 70.0 $\\ 
\  5,000 & $ 67.8 $ & $ 66.7 $ & $ 66.9 $ & $ 70.2 $\\ 
\  8,000 & $ 67.2 $ & $ 68.6 $ & $ 67.8 $ & $ 70.7 $\\ 
\  10,000 & $ \bf 67.9$ & $ \bf68.8 $ & $ \bf 68.3$ & $ \bf 71.0 $\\ 
\hline
\end{tabular}}
\caption{\label{tbl:ablation_data}
Ablation experiments with respect to the amount of unlabeled data. 
}
\end{table}

In this section, we investigate the impact of the amount of unlabeled data on the performance of our method. The datasets used in these experiments are randomly sampled from a total of $10,337$ unlabeled news articles that we have collected, ensuring that they have the same data distribution. \textit{10,000} mentioned in Table \ref{tbl:ablation_data} represents the usage of $10,337$ news articles. 
Compared to our baseline approach where no unlabeled data is utilized, the model trained with an additional $500$ unlabeled news articles does not exhibit improved performance. Since there are only $500$ human-annotated articles available for training, the inclusion of $500$ unlabeled articles with potential noise has a detrimental effect on the model. However, as the amount of unlabeled data increases, the model's performance gradually improves. The performances of models using $8,000$ and $10,000$ articles are similar, suggesting that the performance is approaching saturation.

\end{document}